\documentclass[11pt,a4paper]{article}
\usepackage[hyperref]{emnlp-ijcnlp-2019}
\usepackage{times}
\usepackage{latexsym}
\usepackage{multirow}
\usepackage{url}
\usepackage{amsmath}
\aclfinalcopy 

\title{A Generate-Validate Approach to Answering Questions about Qualitative Relationships}

\author{Arindam Mitra \and Chitta Baral \and Aurgho Bhattacharjee  \and Ishan Shrivastava\\
	Arizona State University\\
	{\tt \{amitra7,chitta,abhatt33,ishrivas\}@asu.edu }}

\date{}

\begin{document}
	\maketitle
	\begin{abstract}
		Qualitative relationships describe how increasing or decreasing one property (e.g. altitude) affects another (e.g. temperature).  They are an important aspect of natural language question answering and are crucial for building chatbots or voice agents where one may enquire about qualitative
		relationships. Recently a dataset about question answering involving qualitative relationships has been proposed, and a few approaches to answer such questions have been explored, in the heart of which lies a semantic parser that converts the natural language input to a suitable logical form. A problem with existing semantic parsers is that they try to directly convert the input sentences to a logical form. Since the output language varies with each application, it forces the semantic parser to learn almost everything from scratch. In this paper, we show that instead of using a semantic parser to produce the logical form, if we apply the generate-validate framework i.e. generate a natural language description of the logical form and validate if the natural language description is followed from the input text, we get a better scope for transfer learning and our method outperforms the state-of-the-art by a large margin of $7.93\%$.
	\end{abstract}
	
	\section{Introduction and Motivation} \label{sec-intro}
	

	The importance of Natural language question answering (NLQA) 
	has greatly accelerated in recent years. It is not only used
	in benchmarking various NLP tasks and their combinations, but some
	NLQA challenges, such as Winograd Schema challenge \cite{levesque2012winograd} and Aristo \cite{clark2015elementary} have been proposed for 
	benchmarking progress in AI as a whole. In terms of applications,
	NLQA plays an important role in human-computer interactions 
	via speech and text and the recent surge in chatbot development, 
	deployment, and usage has further increased  its importance.
	
	In various natural language question answering domains,  applications, and
	challenge corpora one often encounters textual content and questions
	about qualitative relationships. For example, a chatbot developer
	developing a chatbot for a company dealing with windows and curtains
	would need the chatbot to be able to answer questions such as:
	``{\em Will a larger window make the room warmer?}'', and 
	``{\em Will a white curtain in the window make the room cooler?}''. 
	Similarly, in the Aristo \cite{clark2015elementary} corpus there are several items
	that involve qualitative relationships. An example from that corpus is as follows:
	\begin{quote}
		{\em In a large forest with many animals, there are only a small number of bears. 
			Which of these most likely limits the population of bears in the forest? \\
			(A) supply of food\\
			(B) type of tree\\
			(C) predation by carnivores\\
			(D) amount of suitable shelter
		}
	\end{quote}
	
	Considering the importance of being able to answer questions about qualitative relationships in 
	an NLQA setting, recently the QUAREL corpus \cite{tafjord2018quarel} has been proposed. Table \ref{example} shows some examples 
	from the QUAREL corpus.
	\begin{table}[!htb]
		\begin{tabular}{p{200pt}}
			\hline
			I: \em{
				A boomerang thrown into a windy sky heats up quite a bit, but one thrown into a calm sky stays about the same temperature. Which surface puts the least amount of friction on the boomerang? (A) windy sky (B) calm sky
			}
			\\\\

			II: \em{
				Tank the kitten learned from trial and error that carpet is rougher then skin. When he scratches his claws over carpet it generates \_\_\_\_\_\_\_\_ then when he scratches his claws over skin  (A) more heat  (B) less heat 
			}
			\\\\
			
			III: {\em
				The propeller on Kate's boat moved slower in the ocean compared to the river. This means the propeller heated up less in the (A) ocean (B) river
			}
			
			\\
			
			IV: {\em
				Juan is injured in a car accident, which necessitates a hospital stay where he is unable to maintain the strength in his arm. Juan notices that his throwing arm feels extremely frail compared to the level of strength it had when he was healthy. If Juan decides to throw a ball with his friend, when will his throw travel less distance? (A) When Juan's arm is healthy (B) When Juan's arm is weak after the hospital stay.
			}\\\hline
		\end{tabular}
		\caption{Example problems form the QUAREL corpus}
		\label{example}
	\end{table}

	
	Our goal in this paper is to develop a method for answering questions about qualitative relationships, 
	especially with respect to the QUAREL dataset. There are several challenges associated with question 
	answering in this domain. First, it requires reasoning with external knowledge about qualitative relations.
	Although a small knowledge base related to QUAREL has been provided by the QUAREL authors, which we refer to
	as QRKB (Qualitative Relations Knowledge Base),  incorporating that knowledge into the question answering process is a challenge. Second, as pointed out in  \cite{tafjord2018quarel} direct IR based methods, and word association based methods do not do well in this domain. That  is because neither of them properly capture reasoning with external knowledge. A Knowledge Representation  and Reasoning (KR\&R) based approach, that can use reasoning modules from the qualitative reasoning literature \cite{bobrow2012qualitative,weld2013readings} can be employed. For e.g., the problem I from table \ref{example} can be translated to the following tuple: \textit{(qrel(friction, higher, carpet),qrel(heat, higher, carpet),qrel(heat, lower, carpet))}\footnote{This is for illustration purpose.This is not exactly same as the logical form that QUASP or QUASP+ translates to.}. The first component of the tuple \textit{qrel(friction, higher, carpet)} denotes the given fact i.e. ``friction is more on carpet". The second component denotes the claim corresponding to option A i.e. ``more heat is generated on carpet'' and the third component captures the claim corresponding to option B which is ``less heat is generated on carpet''. The reasoning module using the qualitative knowledge that more friction results in more heat can then decide that option A is true. However such approach requires accurate semantic parsing of the text and the question and that is a big challenge. Nevertheless, the authors of QUAREL provide annotations that can facilitate a limited semantic parsing and
	use that  to develop a type constrained neural semantic parser (QUASP) which together with 
	delexicalization results in their best performing system (QUASP+).
	
	Our approach aims to address the drawbacks of using a traditional semantic parser for obtaining the logical representation. Existing semantic parsers are trained to translate the natural language sentences into an application specfic logical representation. Before training, the semantic parsers have some prior knowledge of the input (natural) language, which is normally captured by the word vectors, existing knowledge bases such as WordNet, ConceptNet or parse trees. The target language however is a complete unknown. The model must learn the meaning of the symbols in the target language (i.e. the association between the symbols in the target vocabulary to the ones in input vocabulary) and how to combine these symbols given the input sentence solely from the annotated training data. These expectations naturally increase the demand for more annotated data and these models often suffer if some of the symbols from the output vocabulary do not appear in the training dataset but appear in test set. 
	
	To address these challenges we apply the generate-validate framework \cite{Mitra2019DeclarativeQA} which promotes the following idea: 
	
	\begin{quote}
		If a reasoning algorithm requires facts to be given in a logical form and the application developer has natural language texts at hand, then instead of employing a semantic parser to convert the text to suitable logical facts, \textbf{generate} a natural language description of the logical fact and \textbf{validate} if the text entails the natural language description.
	\end{quote}
	
	Thus instead of generating the logical form from the input problem as is done in \cite{tafjord2018quarel}, we `roughly iterate' over the space of possible logical forms, generate a natural language description for each logical form, validate (score) each of those natural language descriptions using multiple ``textual entailment'' calls and then finally use those scores to detect the correct answer choice. Since, the space of possible logical forms can be quite big, instead of performing a brute-force search we perform an efficient search, which we describe later in section \ref{aproach}.

	Our contributions in this paper are as follows: (1) We show how to apply generate-validate framework to solve the qualitative word problems from QUAREL; (2) We show through experiments that an existing Natural Language Inference dataset, namely SNLI and pre-trained models like BERT can significantly boost the performance on QUAREL when instead of directly generating the logical form, semantic parsing is done through generate-validate. Our method obtains an accuracy of $76.63\%$ which is $7.93\%$ better than QUASP+ model and $20.53\%$ better than QUASP model. We believe that this work will motivate fellow researchers to think differently about semantic parsing and will aid in the development of new models that have a generate-validate architecture at their core and is powered by transfer learning. 
	
	%

	\section{Background}
	
	
	\subsection{The QUAREL dataset}
	The QUAREL dataset \cite{tafjord2018quarel} has 2771 annotated multiple choice story questions. Table \ref{example} shows some sample questions from the QUAREL dataset. Each question in the QUAREL dataset has annotation in the form of logical forms and world literals which we show here for items I and II of Table \ref{example}:\\
	
	\noindent\fbox{%
		\parbox{0.97\linewidth}{%
			{\small
				\noindent \underline{Annotation for Problem I}:\\
				\textbf{Logical Form}\\
				$qval(heat, high, world1), qval(heat, low, world2)\rightarrow$\\ $qrel(friction, lower, world1) ;$\\$ qrel(friction, lower, world2)$\\\\
				\textbf{Literals}\\
				$world1\_literal$ :``windy sky''\\
				$world2\_literal$ : ``calm sky''\\
				
				\noindent \noindent \underline{Annotation for Problem II}:\\
				\textbf{Logical Form}\\
				$qrel(smoothness, lower, world1) \rightarrow$\\
				$qrel(heat, higher, world1) ; qrel(heat, lower, world1)$\\\\
				\textbf{Literals}\\
				$world1\_literal$ : ``carpet''\\ $world2\_literal$: ``skin'
				
	}}}
	\vspace{5pt}
	
	The two examples show two types of logical forms. Syntactically, the logical forms have two parts: the \textit{setup part} that describes the set of explicitly given facts  and the \textit{answer choice part} that gives two claims, one for option A (here after claimA) and another for option B (here after claimB). The setup part and the answer choice part are separated by the `$\rightarrow$' symbol whereas `;' separates the two claims inside the answer choice part.
	
	Both the claims and the given facts are represented by the two predicates, $qrel$ and $qval$.  In the first example the \textit{setup part} provides two facts: $qval(heat, high, world1), qval(heat, low, world2)$ which should be read as: \textit{heat is high in world1} and \textit{heat is low in world2}. The claimA is $qrel(friction, lower, world1)$ which should be read as \textit{friction is lower in world1 \textbf{compared to the other world}} whereas claimB is $qrel(friction, lower, world2)$ which represents \textit{friction is lower in world2 \textbf{compared to the other world}}.  Here, $world1$ and $world2$ are two special symbols which refer to ``windy sky'' and ``calm sky'' respectively. This information is given through the world literal annotation. Each logical form in QUAREL has at max two worlds however the meaning of the worlds i.e. $world1\_literal$ and $world2\_literal$ changes with each problem. Both the predicate $qrel$ and $qval$ has three arguments. The first one is a qualitative property, the second one is called \textit{direction} which could be either low or high and the third one is the special variable \textit{world} which also takes two values $world1$ or $world2$. In this work, we treat $qval$ and $qrel$ uniformly and same natural language description is generated for both of them as there only two worlds and thus the `absolute' ($qval$) and the `relative' ($qrel$) descriptions are equivalent.

	The QRKB of QUAREL has the following 19 qualitative properties: friction, speed, distance, smoothness, heat, 
	loudness, brightness, apparentSize, time, weight, strength, mass, flexibility, exerciseIntensity, acceleration,
	thickness, gravity, breakability, and amountSweat. The QRKB has 25 qualitative relations about 
	pairs of these properties. These relations use the predicates q+ and q-. Some example relations  are:
	q-(friction, speed), and q+(friction, heat). Intuitively, q-(X,Y) means that the amount of X is inversely proportional to the
	amount of Y and q+(X,Y) means that the amount of X is proportional to the amount of Y. Every possible relation pairs are precomputed and stored in QRKB.
	
	
	\subsection{Textual Entailment and NLI} 
	As briefly mentioned in Section~\ref{sec-intro} our approach uses Textual Entailment \cite{dagan2013recognizing} and Natural Language Inference \cite{bowman2015large} models. Natural language inference (NLI) is the task of determining the truth value of a natural language text, called \textit{hypothesis} given another piece of text called \textit{premise}. The list of possible truth values include \textit{entailment}, \textit{contradiction} and \textit{neutral}. \textit{Entailment} means the hypothesis must be true if the premise is true. \textit{Contradiction} indicates that the hypothesis can never be true if the premise is true. \textit{Neutral} pertains to the scenario where the hypothesis can be both true and false as the premise does not provide enough information. Textual Entailment is a binary version of NLI task, where one has to decide if the truth value is \textit{entailment} or not. Table \ref{tab:snli} shows some examples.
	
	Recently, several large scale NLI dataset has been developed. One of which is SNLI \cite{bowman2015large} which we use in this work. Any NLI dataset can be converted to a textual entailment dataset by replacing the \textit{contradiction} and \textit{neutral} label with \textit{not-entailment} label. Among the recent NLI models, the two most popular models are BERT \cite{devlin2018bert} and ESIM \cite{chen2016enhanced} which we use in our implementation.

	\begin{table}[!htb]
		\centering
		\small
		\begin{tabular}{|p{200pt}|}
			\hline
			\textbf{premise:} Tank the kitten learned from trial and error that carpet is rougher then skin. \\
			\textbf{hypothesis:} Carpet is less smooth.\\
			\textbf{label:} entailment.\\
			\hline
			\textbf{premise:} Tank the kitten learned from trial and error that carpet is rougher then skin. \\
			\textbf{hypothesis:} skin is less smooth.\\
			\textbf{label:} not-entailment.\\\hline
		\end{tabular}
		\caption{Example premise-hypothesis pairs with annotated labels.}
		\label{tab:snli}
		\vspace{-10pt}
	\end{table}
	
	

	\section{Proposed approach}
	\label{aproach}
	A qualitative problem $P$ in QUAREL is a sequence of $k$ sentences followed by two option choices. Let $T$ denote the sequence of $k$ sentences and $A_1$ and $A_2$ be the two answer choices. The last sentence in $T$ is a question and is denoted by $Q$. For e.g., for the problem $1$ in Table, $T$ = \textit{A boomerang thrown into a windy sky heats up quite a bit, but one thrown into a calm sky stays about the same temperature. Which surface puts the least amount of friction on the boomerang?}, $A_1$ = \textit{windy sky}, $A_2$ = \textit{calm sky} and $Q$ = \textit{Which surface puts the least amount of friction on the boomerang?} Given such a problem $P = (T,Q,A_1,A_2)$, the task is to decide if $A_1$ is a better answer choice or $A_2$. Our algorithm, namely generate validate qualitative problem solver (gvQPS),  has three key steps, namely \textit{generate, validate} and \textit{inference}, which are discussed in this section.
	
	\medskip \noindent 
	\paragraph{Step 1: Generate}
	Given $T,Q,A_1,$ and  $A_2$ a set $H(T,Q,A_1, A_2)$ of $46\times$$n$ hypothesis such as ``windy sky has more friction'' is created 
	using templates such as ``X has more friction''. Our algorithm uses a total of $46$ manually authored templates. Each template has only one variable $X$ which is substituted by the $n$ noun phrases in the $T$, $Q$, $A_1$ and $A_2$ parts to create the set $H(T,Q,A_1, A_2)$. 
	
	Table \ref{tab:qrt} shows the templates. Each template pertains to a $qrel(P,D, X)$ predicate where $P$ is a qualitative property from QUAREL, $D\in \{low, high\}$, $X$ is a variable representing the textual description of the world. All the properties except \textit{speed} and \textit{distance} have two templates, one for $D=low$ and another for $D=high$. The two properties \textit{speed} and \textit{distance} however have more than two templates to capture different senses.

	For the example 2 from Table \ref{example}, there are a total of $10$ noun-phrases\footnote{according to Spacy constituency parser}, namely  {``heat", ``trial and error", ``claws", ``kitten", ``carpet", ``skin", ``tank kitten", ``error", ``tank", ``trial"}. Thus the set $H(T,Q, A_1, A_2)$ contains a total of $460$ (= $46 \times 10$) hypothesis. Among these the ones related to friction and high are as follows: {\textit{heat has more friction, trial and error has more friction, kitten has more friction, claws has more friction, carpet has more friction, skin has more friction, tank kitten has more friction, error has more friction, tank has more friction, trail has more friction}}.
	
	\paragraph{Step 2: Validate} Recall that the logical form has three parts: the given facts, the claimA and the claimB all of which are represented by the $qrel$ or $qval$ predicate.  In step $1$ the system has generated the set of natural language descriptions of all possible grounded $qval$ predicates, some of which are the given facts, the claimA or claimB. The goal of step $2$ is to precisely identify which statement from $H(T,Q, A_1, A_2)$ is claimA, which statement pertains to claimB and which statements represents the given facts. To do this, the system scores the statements in $H(T,Q, A_1, A_2)$ using two different Textual Entailment functions. Let $given_{score}(.)$, $claimA_{score}(.)$ and $claimB_{score}(.)$ respectively denote the score for a hypothesis to be a given fact, the claimA and the claimB.  These scores are then computed as follows:
	
	\noindent$given_{score}(H_i,T,Q,A_1,A_2) = f_{TE}^{given}(T,H_i)$
	\noindent$$claimA_{score}(H_i,T,Q,A_1,A_2) = f_{TE}^{claim}(QA_1,H_i)$$
	\noindent$$claimB_{score}(H_i,T,Q,A_1,A_2) = f_{TE}^{claim}(QA_2,H_i)$$ 
	
	Here, $QA_1$ and $QA_2$ respectively denotes the concatenation of $Q$,``(option)", $A_1$ and $Q$,``(option)", $A_2$ and $f_{TE}^{given}$ and $f_{TE}^{claim}$ are the two different Textual Entailment functions. $f_{TE}^{given}$ and $f_{TE}^{claim}$ might have same architecture but they are trained on different datasets and take different inputs. For the example II from Table \ref{example} which has a logical representation of $(smoothness, lower, world1) \rightarrow (heat, higher, world1) ; (heat, lower, world1)$, we expect the textual entailment functions to produce the following scores for the sample inputs of table \ref{tab:scoring}. 
	
	\begin{table}[!htb]
		\centering
		\small
		\begin{tabular}{|l|l|}
			\hline
			$given_{score}(``\textit{Carpet is less smooth.}'')$ & 1\\\hline
			$given_{score}(``\textit{Skin is less smooth.}'')$ & 0\\\hline
			$given_{score}(``\textit{Carpet is more smooth.}'')$ & 0\\\hline
			$claimA_{score}(``\textit{Carpet is less smooth.}'')$ & 0\\\hline
			$claimA_{score}(``\textit{more heat is generated on carpet}'')$ & 1\\\hline
			$claimA_{score}(``\textit{less heat is generated on carpet}'')$ & 0\\\hline
			$claimB_{score}(``\textit{more heat is generated on carpet}'')$ & 0\\\hline
			$claimB_{score}(``\textit{less heat is generated on carpet}'')$ & 1\\\hline
			$claimB_{score}(``\textit{less heat is generated on skin}'')$ & 0\\\hline
			
		\end{tabular}
		\caption{Example of expected scores and sample inputs. To save space we do not show the arguments $T,Q,A_1$ and $A_2$ which takes the following value: $T$ = \textit{Tank the kitten learned from trial and error that carpet is rougher then skin. When he scratches his claws over carpet it generates \_\_\_\_\_\_\_\_ then when he scratches his claws over skin}, $Q$ = \textit{When he scratches his claws over carpet it generates \_\_\_\_\_\_\_\_ then when he scratches his claws over skin}, $A_1$ = \textit{more heat, $A_2$} = \textit{less heat}.}
		\label{tab:scoring}
	\end{table}
	
	\begin{table}
		\begin{tabular}{|l|p{120pt}|}
			\hline
			\textbf{(Property, Direction)} & \textbf{Template(s)}\\\hline
			(Friction, high) & X has more friction\\\hline
			(Friction, low) & X has less friction\\\hline
			(Smoothness, high) & X is more smooth\\\hline
			(Smoothness, low) & X is less smooth\\\hline
			(Heat, high) & more heat is generated on X\\\hline
			(Heat, low) & small amount of heat is generated on X\\\hline
			(Loudness, high) & X sounds louder\\\hline
			(Loudness, low) & X sounds softer\\\hline
			(Brightness, high) & X shines more\\\hline
			(Brightness, low) & X looks dim\\\hline
			(apparentSize, high) & X appears big\\\hline
			(apparentSize, low) & X appears small\\\hline
			\multirow{2}{100pt}{(Speed, high)} & X is fast\\\cline{2-2}
			& moves fast through X\\\hline
			\multirow{2}{100pt}{(Speed, low)} & X is slow\\\cline{2-2}
			& moves slowly through X\\\hline
			(time, high) & X takes more time\\\hline
			(time, low) & X takes less time\\\hline
			(weight, high) & X has more weight\\\hline
			(weight, low) & X has less weight\\\hline
			(acceleration, high) & acceleration is more for X\\\hline
			(acceleration, low) & acceleration is less for X\\\hline
			(strength, high) & X has more strength\\\hline
			(strength, low) & X has little strength\\\hline
			\multirow{4}{100pt}{(distance, high)} & travelled more on X\\\cline{2-2}
			&X is far\\\cline{2-2}
			&X travelled more\\\cline{2-2}
			&X threw the object far\\\hline
			\multirow{4}{100pt}{(distance, low)} & travelled less on X\\\cline{2-2}
			&X is near\\\cline{2-2}
			&X travelled less\\\cline{2-2}
			&X could not throw the object far\\\hline
			(thickness, high) & X is thicker\\\hline
			(thickness, low) & X is thin\\\hline
			(mass, high) & X has more mass\\\hline
			(mass, low) & X has less mass\\\hline
			(gravity, high) & X has stronger gravity\\\hline
			(gravity, low) & X has weaker gravity\\\hline
			(flexibility, high) & X is more flexible\\\hline
			(flexibility, low) & X is less flexible\\\hline
			(breakability, high) & X is more likely to break\\\hline
			(breakability, low) & X is less likely to break\\\hline
			(amountSweat, high) &X is exercising more\\\hline
			(amountSweat, low) & X is almost idle\\\hline
			(exerciseIntensity, high) & X is sweating more\\\hline
			(exerciseIntensity, low) & X is sweating less\\\hline
		\end{tabular}
		\caption{Associated templates for each qualitative property.}
		\label{tab:qrt}
	\end{table}

	\paragraph{Step 3: Answer Generation}
	In this step, the system computes the final answer by using the scores that are computed in step $2$. Let $claimA^*$ and $claimB^*$ be the hypothesis in $H(T,Q, A_1, A_2)$ which has respectively the highest $claimA_{score}(.)$ and the highest $claimB_{score}(.)$ score. The answer is option A if $given_{score}(claimA^*)$ is more than $given_{score}(claimB^*)$, otherwise the answer is option B. Here, we assume that the $given_{score}$ will learn to capture the qualitative relationship. For e.g., if it assigns a high score to the hypothesis \textit{skin has less friction}, it will also assign high score to the hypothesis \textit{less heat is generated on skin}.

	\section{Textual Entailment Dataset Generation}
	Our algorithm uses two textual entailment functions namely, $f_{TE}^{given}$ and $f_{TE}^{claim}$ both of which needs to be trained. In this section we describe the process that generates labeled premise-hypothesis pairs from the QUAREL annotations. 
	
	\subsection{Dataset for $f_{TE}^{claim}$} Let $qrel(P^A,D^A,W^A)$ or $qval(P^A,D^A,W^A)$ be the claimA and $qrel(P^B,D^B,W^B)$ or $qval(P^B,D^B,W^B)$ be claimB as per the associated logical form. We use this information to create following annotated premise-hypothesis pairs (we use $1$ to denote \textit{entailment} and $0$ to denote \textit{not-entailment}):
	
	\begin{enumerate}
		\item premise = $QA_1$, hypothesis = generate($P^A,D^A,W^A$) and label = $1$ 
		\item premise = $QA_2$, hypothesis = generate($P^B,D^B,W^B$) and label = $1$
		\item premise = $QA_1$, hypothesis = generate($P^A,opposite(D^A),W^A$) and label = $0$
		\item premise = $QA_2$, hypothesis = generate($P^B,opposite(D^B),W^B$) and label = $0$
		\item If $W_A \ne W_B$, premise = $QA_1$, hypothesis = generate($P^A,D^A,W^B$) and label = $0$
		\item If $W_A \ne W_B$, premise = $QA_2$, hypothesis = generate($P^B,D^B,W^A$) and label = $0$
		\item premise = $QA_1$, hypothesis = generate($P,D,W^A$) and label = $0$ where $P\in QRKB$ and $P\not \in \{P^A,P^B\}$, $D \in \{low,high\}$
		\item premise = $QA_2$, hypothesis = generate($P,D,W^B$) and label = $0$ where $P\in QRKB$ and $P\not \in \{P^A,P^B\}$, $D \in \{low,high\}$
		\item premise = $QA_1$, hypothesis = generate($P^A,D^A,W$) and label = $0$ where $W\in bad$
		\item premise = $QA_2$, hypothesis = generate($P^B,D^B,W$) and label = $0$ where $W\in bad$
		
	\end{enumerate}	
	Here, $generate(.)$ denotes the string that is created for the given input of the type (\textit{qualitative property, direction, world\_literal}) using the templates in table \ref{tab:qrt}; $opposite(D)$ returns the only member  of the set $\{high,low\}\setminus D$ and $bad$ is set of noun phrases from the problem P which does not have any word overlap with either world1\_literal or world2\_literal. For the problem II in table \ref{example}, world1\_literal = ``carpet'' and world1\_literal = ``skin'' and the noun phrases are = {``heat", ``trial and error", ``claws", ``kitten", ``carpet", ``skin", ``tank kitten", ``error", ``tank", ``trial"}. Thus the $bad$ set contain the following elements: {``heat", ``trial and error", ``claws", ``kitten", ``tank kitten", ``error", ``tank", ``trial"}.

	\subsection{Dataset for $f_{TE}^{given}$}
	Similar to $f_{TE}^{claim}$, we create the following annotated premise-hypothesis pairs for each given fact $(P^G,D^G,W^G)$:
	\begin{enumerate}
		\item premise = T, hypothesis = generate($P^G,D^G,W^G$) and label = $1$
		\item premise = T, hypothesis = generate($P^G,opposite(D^G),W^G$) and label = $0$
		\item premise = T, hypothesis = generate($P^G,D^G,\{world1\_literal,\\ world2\_literal\}\setminus W^G$) and label = $0$
		\item premise = T, hypothesis = generate($P^G,D^G,W$) and label = $0$, for all $W \in bad$
		\item premise = T, hypothesis = generate($P,D,W$) and label = $0$, for all property $P$ where none of q+($P,P^A$), q-($P,P^A$),q+($P,P^B$), q-($P,P^B$) is in QRKB, $D$ is either \textit{high} or \textit{low}, $W\in \{world1\_literal, world2\_literal\}$. 
	\end{enumerate}
	However, unlike $f_{TE}^{claim}$, we also create the following annotated premise-hypothesis pairs for each given fact $(P^G,D^G,W^G)$ using QRKB:
	\begin{enumerate}
		\item premise = T, hypothesis = generate($P,D^G,W^G$) and label = $1$, for all property $P$ such that q+($P,P^G$) $in $ QRKB.
		\item premise = T, hypothesis = generate($P,opposite(D^G),W^G$) and label = $1$, for all property $P$ such that q-($P,P^G$) $in $ QRKB.
		\item premise = T, hypothesis = generate($P,D^G,W^G$) and label = $0$, for all property $P$ such that q-($P,P^G$) $in $ QRKB.
		\item premise = T, hypothesis = generate($P,opposite(D^G),W^G$) and label = $0$, for all property $P$ such that q+($P,P^G$) $in $ QRKB.
	\end{enumerate}

	Let $Train^{QUAREL}_{Given}$, $Dev^{QUAREL}_{Given}$ and $Test^{QUAREL}_{Given}$ respectively denote the dataset that are created for $f_{TE}^{given}$ from train, dev and test split of the QUAREL dataset. Similarly, let $Train^{QUAREL}_{Claim}$, $Dev^{QUAREL}_{Claim}$ and $Test^{QUAREL}_{Claim}$  denote the dataset that are created for $f_{TE}^{claim}$ from train, dev and test split of the QUAREL dataset. $Train^{QUAREL}_{Given}$, $Dev^{QUAREL}_{Given}$ and $Test^{QUAREL}_{Given}$ respectively contains $3,58,647$, $50,874$ and $98,057$ premise-hypothesis pairs. On the other hand, $Train^{QUAREL}_{Claim}$, $Dev^{QUAREL}_{Claim}$ and $Test^{QUAREL}_{Claim}$ respectively contains $3,06,545$, $43,914$ and $87,236$ premise-hypothesis pairs. Note that, to make the dataset balanced, the pairs with label $1$ are oversampled. We also use the two-class version of the SNLI dataset to further increase the dataset size.   
	
	\section{Related Work}
	Our work is related to both the works in semantic parsing \cite{zelle1996learning,kwiatkowski2011lexical,berant2013semantic,krishnamurthy2017neural,reddy2014large} and question answering using semantic parsing \cite{lev2004solving,berant2014modeling,Mitra2019DeclarativeQA}. 
	
	The problem of QUAREL is quite similar to the word math problems \cite{hosseini2014learning,kushman2014learning} in the sense that both are story problems and use semantic parsing to translate the input problem to a suitable representation.
	
	Our work is also related to the work in \cite{Mitra2019DeclarativeQA} that uses generate-validate framework to answer questions w.r.t life cycle text. \cite{Mitra2019DeclarativeQA} uses generate-validate framework to verify ``given facts''. Particularly, it shows how rules can be used to infer new information over raw text without using a semantic parser to create a structured knowledge base. The work in \cite{Mitra2019DeclarativeQA} uses a semantic parser to translate the question into one of the predefined forms. In our work, however we use generate-validate for both question and ``given fact" understanding.
	
	The work of \cite{tafjord2018quarel} is most related to us. \cite{tafjord2018quarel} proposes two models for QUAREL. One uses a state-of-the-art semantic parser \cite{krishnamurthy2017neural} to convert the input problem to the desired logical representation. They call this model QUASP, which obtains an accuracy of $56.1\%$. The other model, called QUASP+ uses a delexicalization step before giving the input to the semantic parser. The delexicalization step identifies the value(s) of world1\_literal and word2\_literal and then replaces all the occurrences of those strings in the text by the symbol ``world1" and ``world2''. The modified input is then passed to the semantic parser. The delexicalization helps the semantic parser by giving explicit pointers to world1 and world2, which results in an accuracy of 68.7\%. Our model does not use such preprocessing and still performs significantly better than QUASP+ model.

	\section{Experimental Evaluation}
	We use the notation $f^M_D$ to denote that the textual entailment model in use is M which can be either ESIM or BERT and the model M is trained on the dataset D which can be any of following: $Train_{Given}^{QUAREL}$, $Train_{Given}^{QUAREL} \cup Train^{SNLI}$, $Train_{Claim}^{QUAREL}$, $Train_{Claim}^{QUAREL} \cup Train^{SNLI}$. Correspondingly there are a total of $4$ possible values for $f^{given}_{TE}$ namely $f^{ESIM}_{Train_{Fact}^{QUAREL}}$, $f^{BERT}_{Train_{Given}^{QUAREL}}$, $f^{ESIM}_{Train_{Given}^{QUAREL} \cup Train^{SNLI}}$ and $f^{BERT}_{Train_{Fact}^{QUAREL} \cup Train^{SNLI}}$. Similarly, there are a total of $4$ possible values for $f^{claim}_{TE}$ namely $f^{ESIM}_{Train_{claim}^{QUAREL}}$, $f^{BERT}_{Train_{claim}^{QUAREL}}$, $f^{ESIM}_{Train_{claim}^{QUAREL} \cup Train^{SNLI}}$ and $f^{BERT}_{Train_{Fact}^{QUAREL} \cup Train^{SNLI}}$. Table \ref{exp1} shows the results of our algorithm for all these $4\times 4$= $16$ combinations.
	
	\begin{table}
		\begin{tabular}{|l|l|l|l|}
			\hline
			$f^{given}_{TE}$ & $f^{claim}_{TE}$ & Dev(\%) &Test(\%) \\\hline
			$f^{ESIM}_{G_1}$ & $f^{ESIM}_{C_1}$ &67.27&71.2\\\hline
			$f^{ESIM}_{G_1}$ & $f^{BERT}_{C_1}$ &62.23&69.12\\\hline
			$f^{ESIM}_{G_1}$ & $f^{ESIM}_{C_2}$&66.54&69.57\\\hline
			$f^{ESIM}_{G_1}$ & $f^{BERT}_{C_2}$&59.71& 67.39\\\hline
			
			$f^{BERT}_{G_1}$ & $f^{ESIM}_{C_1}$ &67.99& 71.56\\\hline
			$f^{BERT}_{G_1}$ & $f^{BERT}_{C_1}$ &67.62 &69.38\\\hline
			$f^{BERT}_{G_1}$ & $f^{ESIM}_{C_2}$&62.95 &69.2\\\hline
			$f^{BERT}_{G_1}$ & $f^{BERT}_{C_2}$& 68.35&67.93\\\hline

			$f^{ESIM}_{G_2}$& $f^{ESIM}_{C_1}$& 68.34 &67.21\\\hline
			$f^{ESIM}_{G_2}$& $f^{BERT}_{C_1}$ &59.35 &66.49\\\hline
			$f^{ESIM}_{G_2}$& $f^{ESIM}_{C_2}$&66.55 &66.3\\\hline
			$f^{ESIM}_{G_2}$& $f^{BERT}_{C_2}$&58.63 &64.3\\\hline
			
			$f^{BERT}_{G_2}$& $f^{ESIM}_{C_1}$ &\textbf{73.38} &\textbf{76.63}\\\hline
			$f^{BERT}_{G_2}$& $f^{BERT}_{C_1}$ & 72.66 &75.36\\\hline
			$f^{BERT}_{G_2}$& $f^{ESIM}_{C_2}$&70.50 &73.55\\\hline
			$f^{BERT}_{G_2}$& $f^{BERT}_{C_2}$& 73.02&70.29\\\hline
		\end{tabular}
	\caption{shows the accuracy on dev and test set of QUAREL for various choice of $f^{given}_{TE}$ and $f^{claim}_{TE}$. Here, $G_1,G_2,C_1$ and $C_2$ respectively represents $Train_{Given}^{QUAREL}$, $Train_{Given}^{QUAREL} \cup Train^{SNLI}$, $Train_{Claim}^{QUAREL}$, $Train_{Claim}^{QUAREL} \cup Train^{SNLI}$.}
	\label{exp1}
	\end{table}
	
	\begin{itemize}
		\item The best performance is achieved when, $f^{BERT}_{Train_{Fact}^{QUAREL} \cup Train^{SNLI}}$ is used as $f^{given}_{TE}$ and $f^{ESIM}_{Train_{claim}^{QUAREL}}$ is used as $f^{claim}_{TE}$. We refer to this as gvQPS\textsuperscript{B\textsuperscript{+}E}. The performance of this combination is $5.07\%$ more than the combination of $f^{ESIM}_{Train_{Fact}^{ESIM}}$ and $f^{ESIM}_{Train_{claim}^{QUAREL}}$ which shows the boost offered by BERT and SNLI.
		\item The accuracy normally drops when SNLI dataset is used in the training for the $f^{claim}_{TE}$ function irrespective of the model on both dev and test set. We speculate that this happens because the premise in SNLI contain proper sentences whereas the premise in the $Train_{claim}^{QUAREL}$ are options appended to questions and thus have different distributions.
		\item ESIM models perform consistently better as $Train_{claim}^{QUAREL}$ than BERT models irrespective of the training dataset on both dev and test set.
	\end{itemize}
	
	Table \ref{exp2} compares our best performing method with other approaches. As shown, in table \ref{exp2} our model provides an improvement of $7.93\%$ over the previous state-of-the-art QUASP+.
	\begin{table}[!htb]
		\begin{tabular}{|l|l|}
			\hline
			Model & Accuracy(\%) \\\hline
			IR & 48.6\\\hline
			PMI & 50.5\\\hline
			QUASP & 56.1\\\hline
			QUASP+ & 68.7\\\hline
			\textbf{gvQPS\textsuperscript{B\textsuperscript{+}E}} & 76.63\\\hline
		\end{tabular}
	\caption{Comparing our best performing model with existing solvers of QUAREL.}
	\label{exp2}

	\end{table}
	
	\paragraph{Error Analysis}
	Our best model, gvQPS\textsuperscript{B\textsuperscript{+}E} fails to solve $129$ problems. The majority of the error occurs due to the error in $given_{score}(.)$. The following figure shows two examples of error with $claimA^*$ and $claimB^*$ and the scores of the relevant hypothesis by $given_{score}(.)$.
	
		\noindent\fbox{%
		\parbox{0.97\linewidth}{%
			{\small
				\noindent \underline{Error Example I}:\\
				\textit{Nell has very thick hair; Lynn's hair is much thinner. Whose hair is stronger? (A) Nell (B) Lynn}\\
				$claimA^*$ : ($strength, high$, `Nell')\\
				$claimB^*$ : ($strength, high$, `Lynn's hair')\\
				\textbf{Sample $given_{score}(.)$ scores}\\
				lynn 's hair has more strength, $0.01$\\
				nell has more strength, $0.00003$\\\\
				
				\noindent \underline{Error Example II}:\\
				\textit{David noticed that it was harder to push his snow blower on snowy pavement than on dry pavement. This is because the dry pavement has (A) more friction or (B) less friction}\\
				$claimA^*$ : ($friction, high$, `dry pavement')\\
				$claimB^*$ : ($strength, low$, `dry pavement')\\
				\textbf{Sample $given_{score}(.)$ scores}\\
				dry pavement has more friction, $0.9645242997992661$\\
				dry pavement has less friction, $0.000003$
	}}}
	\vspace{5pt}
	
	As seen in the above figure, for both the error examples, the $claimA^*$ and $claimB^*$ have been identified correctly, however the $given_{score}(.)$ predicts wrongly which results in an error. 
	
	\section{Conclusion}
	Semantic Parsing has been quite useful in solving problems that require sophisticated reasoning such as math word problems, logic puzzles, qualitative word problems, question-answering over database or query understanding and has been extensively used in many applications. However, traditional semantic parser has certain drawbacks which can be potentially addressed with the generate-validate framework. In this work, we have shown how to successfully apply the generate-validate framework to solve qualitative word problems and have shown the opportunities of transfer learning that are available in this framework. Our future work is to apply the generate-validate framework to other applications which use semantic parsing. Our work also connects the popular task of Natural Language Inference to the applications of semantic parsing and any improvements in the Natural Language Inference models will naturally improve the performance of our models. 
	
	\bibliography{quarel}
	\bibliographystyle{acl_natbib}
	
\end{document}